\documentclass[10pt,twocolumn,letterpaper]{article}

\usepackage[pagenumbers]{cvpr} 

\usepackage[utf8]{inputenc} 
\usepackage[T1]{fontenc}    
\usepackage{url}            
\usepackage{booktabs}       
\usepackage{amsfonts}       
\usepackage{amssymb}
\usepackage{amsmath}
\usepackage{nicefrac}       
\usepackage{microtype}      
\usepackage[dvipsnames,table,xcdraw]{xcolor}        
\usepackage{array}
\usepackage{graphicx}
\usepackage{subcaption}
\usepackage{mwe}
\usepackage[percent]{overpic}
\usepackage{multirow}
\usepackage{algorithmic}
\usepackage{algorithm}
\usepackage{verbatim} 

\usepackage[pagebackref,breaklinks,colorlinks]{hyperref}

\makeatletter
\newcommand{\HEADER}[1]{\ALC@it\underline{\textsc{#1}}\begin{ALC@g}}
\newcommand{\ENDHEADER}{\end{ALC@g}}
\makeatother

\usepackage[capitalize]{cleveref}
\crefname{section}{Sec.}{Secs.}
\Crefname{section}{Section}{Sections}
\Crefname{table}{Table}{Tables}
\crefname{table}{Tab.}{Tabs.}

\newcolumntype{L}[1]{>{\raggedright\let\newline\\\arraybackslash\hspace{0pt}}m{#1}}
\newcolumntype{C}[1]{>{\centering\let\newline\\\arraybackslash\hspace{0pt}}m{#1}}
\newcolumntype{R}[1]{>{\raggedleft\let\newline\\\arraybackslash\hspace{0pt}}m{#1}}

\begin{document}

\title{PubTables-1M: Towards comprehensive table extraction from unstructured documents}

\author{%
  Brandon Smock\qquad Rohith Pesala\qquad Robin Abraham\\
  Microsoft\\
  Redmond, WA \\
  {\tt\small brsmock,ropesala,robin.abraham@microsoft.com} \\
}

\maketitle

\begin{abstract}
Recently, significant progress has been made applying machine learning to the problem of table structure inference and extraction from unstructured documents.
However, one of the greatest challenges remains the creation of datasets with complete, unambiguous ground truth at scale.
To address this, we develop a new, more comprehensive dataset for table extraction, called PubTables-1M.
PubTables-1M contains nearly one million tables from scientific articles, supports multiple input modalities, and contains detailed header and location information for table structures, making it useful for a wide variety of modeling approaches.
It also addresses a significant source of ground truth inconsistency observed in prior datasets called oversegmentation, using a novel \emph{canonicalization} procedure.
We demonstrate that these improvements lead to a significant increase in training performance and a more reliable estimate of model performance at evaluation for table structure recognition.
Further, we show that transformer-based object detection models trained on PubTables-1M produce excellent results for all three tasks of detection, structure recognition, and functional analysis without the need for any special customization for these tasks.
Data and code will be released at \url{https://github.com/microsoft/table-transformer}.
\end{abstract}

\section{Introduction}

\begin{figure}[]
    \centering
    \includegraphics[height=2.8cm]{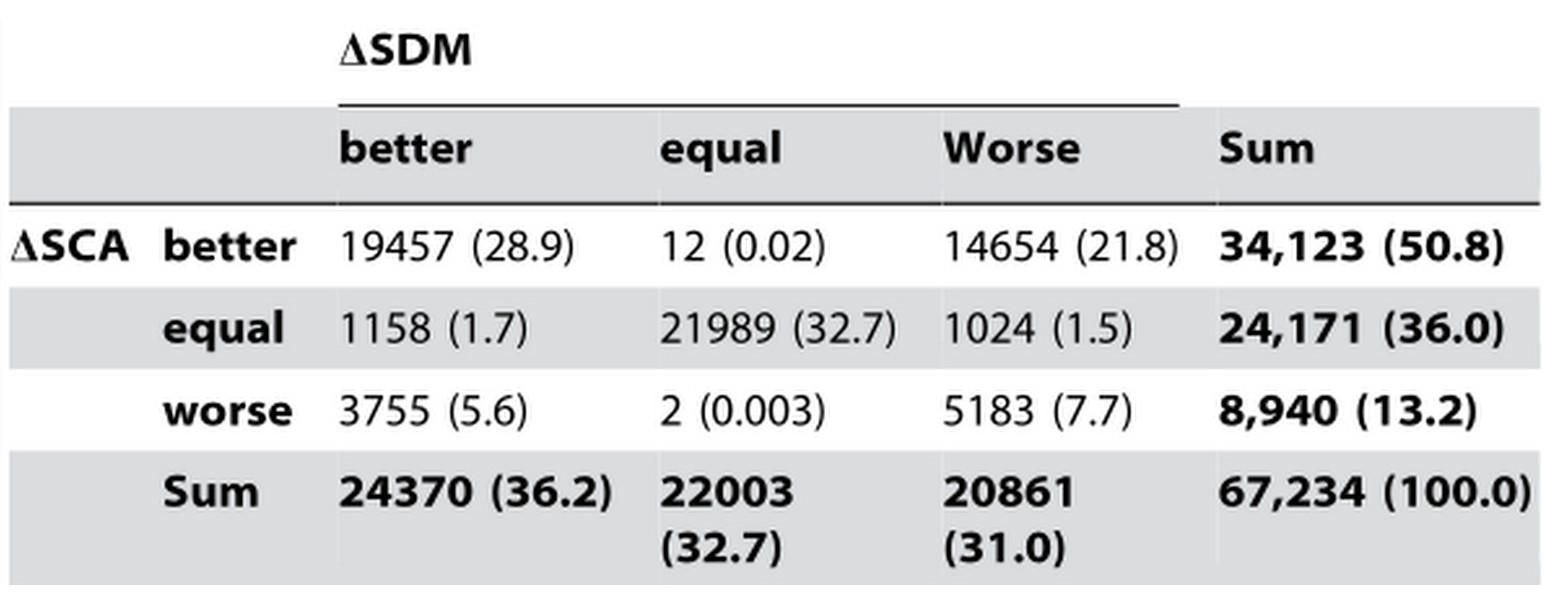}
    \caption{An example of a presentation table whose underlying structure must be inferred, either manually or by automated systems.}
    \label{fig:example1}
\end{figure}

A table is a compact, structured representation for storing data and communicating it in documents and other manners of presentation.
In its presented form, however, a table, such as the one in \cref{fig:example1}, may not and often does not explicitly represent its logical structure.
This is an important problem as a significant amount of data is communicated through documents, but without structure information this data cannot be used in further applications.

\begin{figure*}[]
  \centering
  \includegraphics[height=6.8cm]{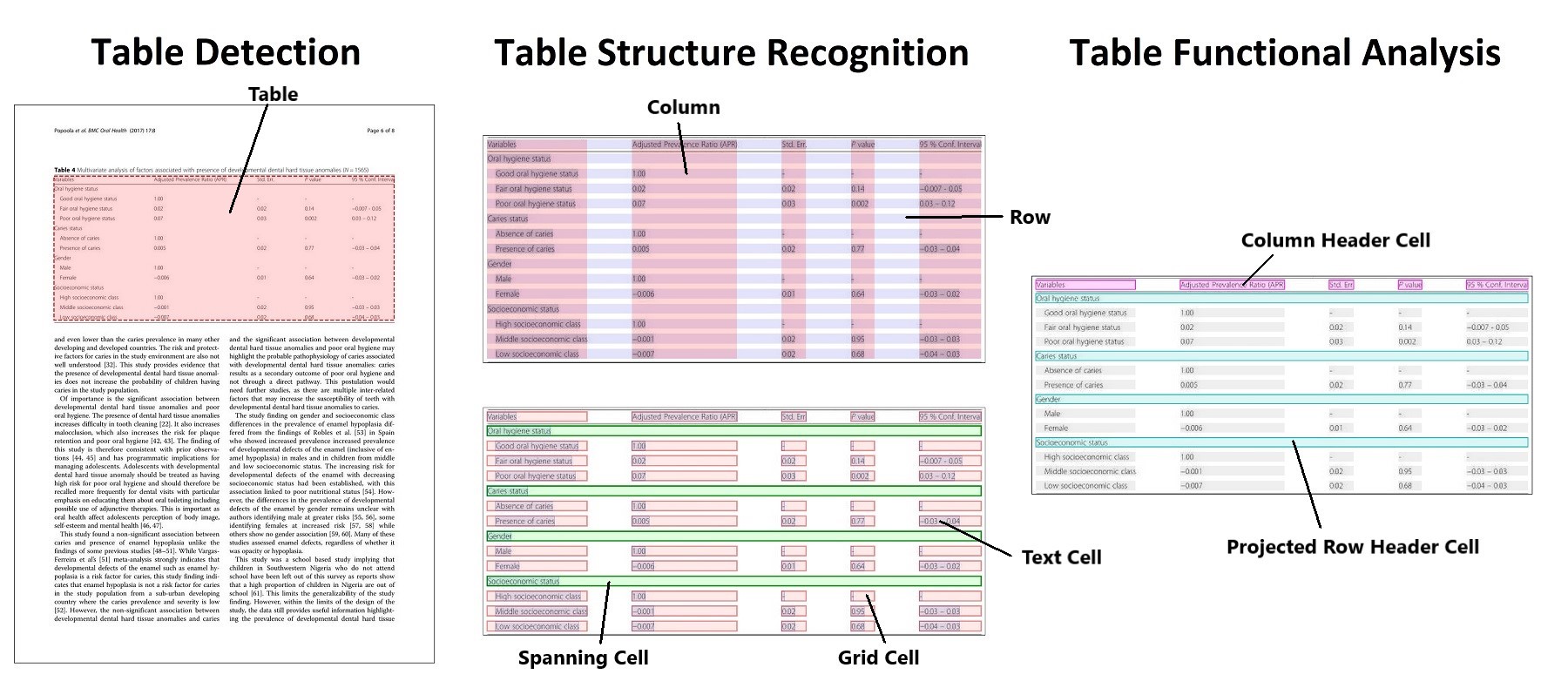}
  \caption{Illustration of the three subtasks of table extraction addressed by the PubTables-1M dataset.}
  \label{fig:table_extraction_tasks}
\end{figure*}

The problem of inferring a table's structure from its presentation and converting it to a structured form is known as table extraction (TE).
TE entails three subtasks \cite{gobel2013icdar}, which we illustrate in \cref{fig:table_extraction_tasks}: \textit{table detection} (TD), which locates the table; \textit{table structure recognition} (TSR), which recognizes the structure of a table in terms of rows, columns, and cells; and \textit{functional analysis} (FA), which recognizes the keys and values of the table.
TE is challenging for automated systems \cite{schreiber2017deepdesrt, zhong2019image, li2020tablebank, paliwal2019tablenet} due to the wide variety of formats, styles, and structures found in presented tables.

Recently, there has been a shift in the research literature from traditional rule-based methods \cite{gatterbauer2007towards, oro2009trex, shigarov2015table} for TE to data-driven methods based on deep learning (DL) \cite{schreiber2017deepdesrt, prasad2020cascadetabnet, zheng2021global}.
The primary advantage of DL methods is that they can learn to be more robust to the wide variety of table presentation formats.
However, manually annotating tables for TSR is a difficult and time-consuming process \cite{hu2001table}.
To overcome this, researchers have turned recently to crowd-sourcing to construct larger datasets \cite{li2020tablebank, zhong2019image, zheng2021global}.
These datasets are assembled from tables appearing in documents created by thousands of authors, where an annotation for each table's structure and content is available in a markup format such as HTML, XML, or LaTeX.

While crowd-sourcing solves the problem of dataset size, repurposing annotations originally unintended for TE and automatically converting these to ground truth presents its own set of challenges with respect to completeness, consistency, and quality.
This includes not only what information is present but how \emph{explicitly} this information is represented.
For instance, markup annotations do not encode spatial coordinates for cells, and they only encode logical relationships implicitly through cues such as layout \cite{tengli2004learning}.
Not only does this lack of explicit information limit the range of supervised modeling approaches, it also limits the quality control that can be done to verify the annotations' correctness.

\begin{figure*}[]
  \centering
  \begin{subfigure}[b]{0.49\linewidth}
	\centering
	\includegraphics[height=2.8cm]{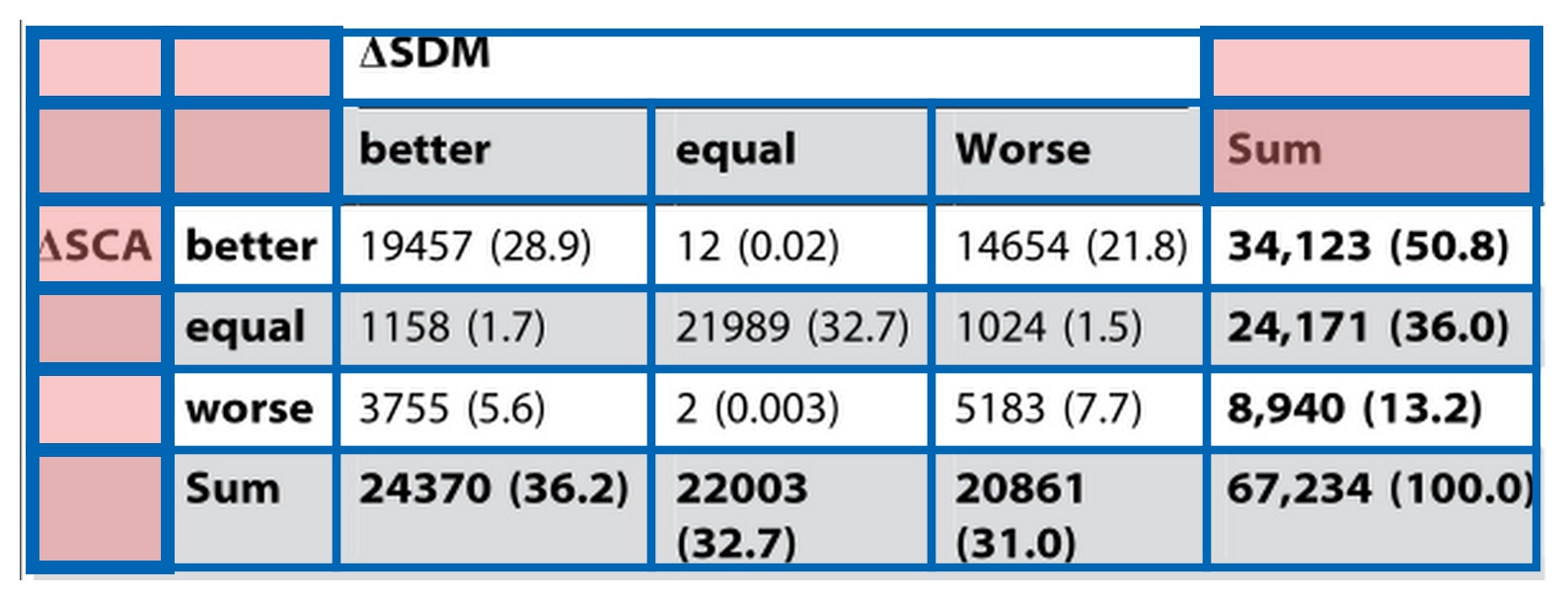}
    \caption{Oversegmented structure annotation}
    \label{subfig:oversegmentation.1}
  \end{subfigure}
  \begin{subfigure}[b]{0.49\linewidth}
	\centering
	\includegraphics[height=2.8cm]{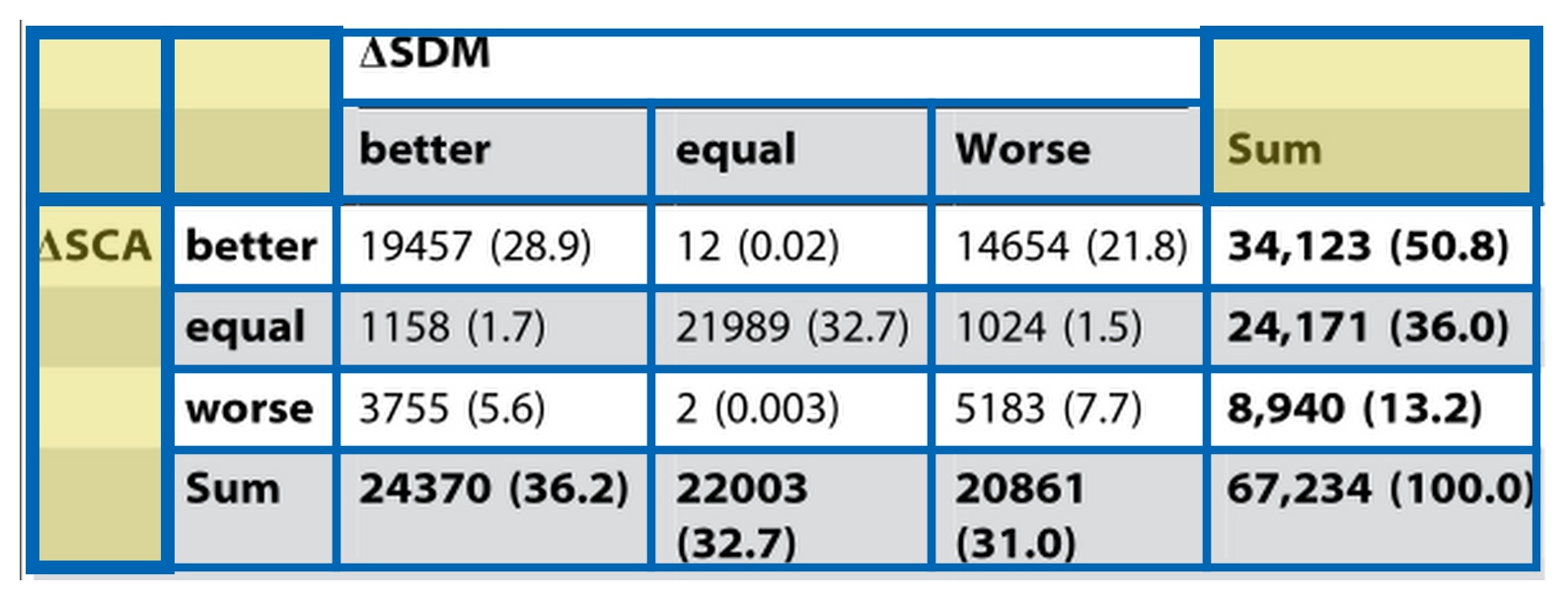}
    \caption{Canonical structure annotation}
    \label{subfig:oversegmentation.2}
  \end{subfigure}
  \caption{In the above example, the structure annotation on the left is \emph{oversegmented}, creating extra blank cells in the headers. The canonical structure annotation on the right merges these cells and captures its true logical structure.}
  \label{fig:oversegmentation}
\end{figure*}

Another significant challenge for the use of crowd-sourced annotations is that these structure annotations encoded in markup often exhibit an issue we refer to as \emph{oversegmentation}.
Oversegmentation occurs in a structure annotation when a spanning cell in a header is split into multiple grid cells.
We give examples of this in \cref{fig:oversegmentation}.
Oversegmentation in markup usually has no effect on how a table appears due to borders between cells being invisible, leaving a presentation table's implicit logical structure and interpretation unaffected.
However, oversegmentation can lead to significant issues when used as ground truth for model training and evaluation.

The first issue is that an oversegmented annotation contradicts the logical interpretation of the table that its presentation is meant to suggest.
For instance, oversegmenting a cell annotation may indicate that its text applies to only one row when its presentation form suggests its text is meant to apply to several rows, as in the cell in column 1, row 3 in \cref{fig:oversegmentation}.
This is problematic for use as ground truth to teach a machine learning model to correctly interpret a table's structure.
Even if oversegmented annotations were considered a valid intepretation of a table's structure, allowing them would and does lead to ambiguous and inconsistent ground truth, due to there then being multiple possible valid interpretations for a table's structure, such as in \cref{fig:oversegmentation}.
This violates the standard modeling assumption that there is exactly one correct ground truth annotation for each table.
Thus, datasets that contain oversegmented annotations in them lead to inconsistent, contradictory feedback during training and an underestimate of true performance during evaluation.

To address these and other challenges, we develop a new large-scale dataset for table extraction called PubTables-1M.
PubTables-1M contains nearly one million tables from scientific articles in the PubMed Central Open Access\footnote{https://www.ncbi.nlm.nih.gov/pmc/tools/openftlist/} (PMCOA) database.
Among our contributions:
\begin{itemize}
	\item PubTables-1M is nearly twice as large as the current largest comparable dataset and addresses all three tasks of table detection (TD), table structure recognition (TSR), and functional analysis (FA).
	\item Compared to prior datasets, PubTables-1M contains richer annotation information, including annotations for projected row headers and bounding boxes for all rows, columns, and cells, including blank cells. It also includes annotations on their original source documents, which supports multiple input modalities and enables a wide range of potential model architectures.
	\item We introduce a novel \textit{canonicalization} procedure that corrects oversegmentation and whose goal is to ensure each table has a unique, unambiguous structure interpretation.
	\item To reduce additional sources of error, we implement several quality verification and control steps and provide measurable guarantees about the quality of the ground truth.
	\item We show that data improvements alone lead to a significant increase in performance for TSR models, due both to improved training and a more reliable estimate of performance at evaluation.
	\item Finally, we apply the Detection Transformer (DETR) \cite{carion2020end} for the first time to the tasks of TD, TSR, and FA, and demonstrate how with PubTables-1M all three tasks can be addressed with a transformer-based object detection framework without any special customization for these tasks.
\end{itemize}

\section{Related Work}

\paragraph{Structure recognition datasets} The first dataset to address all three table extraction tasks was the ICDAR-2013 dataset \cite{gobel2013icdar}.
It remains popular for benchmarking TSR models due to its quality and relative completeness compared to other datasets.
However, as a source of training data for table extraction models it is limited, containing only 248 tables for TD and TSR and 92 tables for FA.

Recently, larger datasets \cite{chi2019complicated, zhong2019image, li2020tablebank, zheng2021global} have been created by collecting crowd-sourced table annotations automatically from existing documents.
We summarize these datasets in \cref{tab:tsr}.
Each source table has an annotation for its content and structure in a markup format such as HTML, XML, or LaTeX.
Various methods are used to determine each table's spatial location within its containing document to create a correspondence between its markup and its presentation.
From there, datasets commonly frame the TSR task as: given an input table, output its cell structure---the assignment of cells to rows and columns---and the text content for each cell, with image and HTML being example input and output formats, respectively, for these.

\begin{table*}
  \footnotesize
  \caption{Comparison of crowd-sourced datasets for table structure recognition.}
  \label{tab:tsr}
  \centering
  \begin{tabular}{L{0.16\textwidth}L{0.07\textwidth}R{0.06\textwidth}C{0.07\textwidth}C{0.05\textwidth}C{0.07\textwidth}C{0.11\textwidth}C{0.08\textwidth}}
    \toprule
    \textbf{Dataset} & \textbf{Input Modality} & \textbf{\# Tables} & \textbf{Cell \mbox{Topology}} & \textbf{Cell \mbox{Content}} & \textbf{Cell \mbox{Location}} & \textbf{Row \& Column Location} & \textbf{Canonical Structure} \\
    \midrule
    TableBank \cite{li2020tablebank} & Image & 145K & \checkmark & & & &\\
    SciTSR \cite{chi2019complicated} & $\textrm{PDF}^*$ & 15K & \checkmark & \checkmark & & & \\
    PubTabNet \cite{zhong2019image, zheng2021global} & Image & $\textrm{510K}^\ddag$ & \checkmark & \checkmark & $\checkmark^\dagger$ & & \\
    FinTabNet \cite{zheng2021global} & $\textrm{PDF}^*$ & 113K & \checkmark & \checkmark & $\checkmark^\dagger$ & & \\
    \midrule
    \textbf{PubTables-1M (ours)} & $\textbf{PDF}^*$ & \textbf{948K} & \textbf{\checkmark} & \textbf{\checkmark} & \textbf{\checkmark} & \textbf{\checkmark} & \textbf{\checkmark} \\
    \midrule
    \multicolumn{7}{l}{\scriptsize $^*$Multiple input modalities, such as image or text, can be derived from annotated PDF data.} \\
    \multicolumn{7}{l}{\scriptsize $^\ddag$The authors release annotations for 510K of the 568K total tables in their dataset.} \\
    \multicolumn{7}{l}{\scriptsize $^\dagger$For these datasets, cell bounding boxes are given for non-blank cells only and exclude any non-text portion of a cell.} \\
  \end{tabular}
\end{table*}

More recently, two large datasets, FinTabNet and an enhanced version of PubTabNet, have added location information for cells, similar to ICDAR-2013.
Adding location information enables the TSR task to be framed as outputting cell location instead of cell content, with cell content extraction being a trivial subsequent step.
This increases the range of possible supervised modeling approaches.
However, the bounding boxes for cells defined by these datasets cover only the text portion of each and exclude any additional whitespace a cell might contain.
This has a few implications, such as making bounding boxes for blank cells undefined and excluding attributes contributed by whitespace, such as text indentation and alignment.
Therefore, one question left open by prior work is how to define bounding boxes for all cells, including blank cells.

There are additional challenges related to annotation completeness and quality that have not been addressed by prior datasets.
In terms of completeness, prior large-scale datasets have also not included bounding boxes for rows and columns.
Additionally, most datasets do not annotate the column header, and no prior large-scale dataset exists that specifies the row header of a table.
This not only limits the range of modeling approaches that can be applied to TSR but limits how completely the overall TE task can be solved.

Another open challenge is automated verfication and measurement of annotation quality, which is important due to the impracticality of verifying large-scale annotations manually.
Prior datasets have also not addressed the significant issue of oversegmented annotations.
These are important issues, as noise and mistakes in training data potentially harm learning and in evaluation data potentially lead to an underestimate of model performance.
But currently the extent to which these issues affect model training and evaluation is unexplored.

\paragraph{Modeling approaches}

One of the most common modeling approaches for TSR is to frame the task as some form of object detection \cite{schreiber2017deepdesrt, prasad2020cascadetabnet, zheng2021global}.
Other approaches include those based on image-to-text \cite{li2020tablebank} and graph-based approaches \cite{qasim2019rethinking, chi2019complicated}.
While a number of general-purpose architectures, such as Faster R-CNN \cite{ren2015faster}, exist for these model patterns, the unique characteristics of table and the relative lack of training data have both contributed to the commonly observed underperformance of these models when applied to TSR out-of-the-box.

To get around deficiencies in training data, some approaches model TSR in ways that are only partial solutions to the task, such as row and column detection in DeepDeSRT \cite{schreiber2017deepdesrt}, which ignores spanning cells, or image-to-markup without cell text content, as in models trained on TableBank \cite{li2020tablebank}.
Other approaches use custom pipelines that branch to consider different cases separately, such as training separate models to recognize tables with and without visible borders surrounding every cell \cite{prasad2020cascadetabnet, zheng2021global}.
Many of the previously mentioned approaches also use engineered model components or custom training procedures, and incorporate rules or other unlearned processing stages tailored to the TSR task, which brings in prior knowledge to lessen the burden placed on learning the task from data.
Currently, no solution exists that uses a simple supervised learning approach with an off-the-shelf architecture, solves the TSR task completely, and achieves state-of-the-art performance.

\section{PubTables-1M}\label{sec:dataset}

In this section, we describe the process used to develop PubTables-1M.
First, to obtain a large source of annotated tables, we choose the PMCOA corpus, which consists of millions of publicly available scientific articles.
In the PMCOA corpus, each scientific article is given in two forms: as a PDF document, which visually presents the article, and as an XML document, which provides a semantic description and hierarchical organization of the document's elements.
Each table's content and structure is specified using standard HTML tags.

However, because this data was not intended for use as ground truth for table extraction modeling, it does not explicitly label or guarantee multiple things that would be helpful for this purpose.
For instance, although the same tables appear in both documents, no direct correspondence between them is given, nor the spatial location of each table.
In terms of data quality, while tables are generally annotated reliably, it is not guaranteed that column headers are annotated completely or that text content as annotated exactly matches the text content as it appears in the PDF.
Finally, some labels, such as the row header for each table, are not annotated at all.

The basic approach we take to overcome these issues is first we attempt to reliably infer as much missing annotation information as possible (for instance, the spatial location of each table) from the information that is present, then we verify that each annotation meets certain requirements for consistency.
In some cases, we correct an annotation to attempt to make it more consistent, such as merging cells that are oversegmented.
We consider certain requirements for tables to be strict and samples whose annotations violate these are removed.
This provides a set of conditions for quality and consistency that the annotations are guaranteed to meet.
In the rest of this section, we describe these conditions and the steps we take to derive ground truth that meets them.

\paragraph{Alignment}
Text in a PDF document has spatial location $[x_\text{min}, y_\text{min}, x_\text{max}, y_\text{max}]$, while text in an XML document appears inside semantically labeled tags.
Because the correspondence between these is not given, the first step in creating PubTables-1M is to match the text content from both.
We process the PDF document into a sequence of characters each with their associated bounding box and use the Needleman-Wunsch algorithm \cite{needleman1970general} to align this with the character sequence for the text extracted from each table HTML.
This connects the text within each HTML tag to its spatial location with the PDF document.
For each cell with text, we compute the union of the bounding boxes for each character of the cell's text, which we refer to as a \emph{text cell} bounding box.

\paragraph{Completion}
Following alignment, we complete the spatial annotations to define bounding boxes for rows, columns, and the entire table.
The bounding box for the table is defined simply as the union of all text cell bounding boxes.
The $x_\text{min}$ and $x_\text{max}$ of the bounding box for each row are defined as the $x_\text{min}$ and $x_\text{max}$ of the table, giving every row the same horizontal length.
The $y_\text{min}$ and $y_\text{max}$ of the bounding box for each row, $m$, are defined as the $y_\text{min}$ and $y_\text{max}$ of the union of the text cells for each cell whose starting row or ending row is $m$.
Similarly, the $y_\text{min}$ and $y_\text{max}$ of the bounding box for each column are defined as the $y_\text{min}$ and $y_\text{max}$ of the table.
The $x_\text{min}$ and $x_\text{max}$ of the bounding box for each column, $n$, are defined as the $x_\text{min}$ and $x_\text{max}$ of the union of the text cell for each cell whose starting column or ending column is $n$.
From these definitions, the \emph{grid cell} for each cell is defined as the union of the bounding boxes of the cell's rows intersected with the union of the bounding boxes for its columns.
Unlike the text cell, the grid cell is defined even for blank cells.

\paragraph{Canonicalization}\label{par:canonicalization}
The primary goal of the canonicalization step is to correct oversegmentation in a table's structure annotations.
To do this, we need to make assumptions about a table's intended structure.
As the canonicalization algorithm itself is relatively simple, we first describe it, then detail the assumptions that motivate it.
Put simply, canonicalization amounts to merging adjacent cells under certain conditions.
This algorithm is given in \cref{alg:canonicalization}.
But because it only operates on cells in the headers, HTML does not have a tag for specifying a table's row header, and we observed that the column headers for tables in the PMCOA corpus are not always correct, we also include steps for inferring additional header cells that we believe can be reliably inferred in PMCOA markup annotations.
These additional steps significantly increase the number of cells whose oversegmentation we are able to correct.

\begin{algorithm}
  \caption{PubTables-1M Canonicalization}
  \label{alg:canonicalization}
  \small
  \begin{algorithmic}[1]
    \HEADER{Add cells to the column and row headers}
		\STATE Split every blank spanning cell into blank grid cells \label{algline:splitblank}
		\STATE \textbf{if} the first row starts with a blank cell \textbf{then} add the first row to the column header
		\IF {there is at least one row labeled as part of the column header}
			\STATE \textbf{while} every column in the column header does not have at least one complete cell that only spans that column \textbf{do:} add the next row to the column header
		\ENDIF
		\STATE \textbf{for each} row \textbf{do:} \textbf{if} the row is not in the column header and has exactly one non-blank cell that occupies the first column \textbf{then} label it a projected row header \label{algline:prh}
		\STATE \textbf{if} any cell in the first column below the column header is a spanning cell or blank \textbf{then} add the column (below the column header) to the row header
    \ENDHEADER
    \HEADER{Merge cells}
		\STATE \textbf{for each} cell in the column header \textbf{do} recursively merge the cell with any adjacent cells above and below in the column header that span the exact same columns \label{algline:mergecolumn}	
		\STATE \textbf{for each} cell in the column header \textbf{do} recursively merge the cell with any adjacent blank cells below it if every adjacent cell below it is blank and in the column header \label{algline:mergedown}
		\STATE \textbf{for each} cell in the column header \textbf{do} recursively merge the cell with any adjacent blank cells above it if every adjacent cell above it is blank
		\STATE \textbf{for each} projected row header \textbf{do} merge all of the cells in the row into a single cell
		\STATE \textbf{for each} cell in the row header \textbf{do} recursively merge the cell with any adjacent blank cells below it \label{algline: rowheadermerging}
    \ENDHEADER
\end{algorithmic}
\end{algorithm}

We first assume that each table has an intended structure consistent with the Wang model\cite{wang1996tabular}, which in a study Wang found was true for 97 percent of observed tables.
Under this model, the headers of the table each have a hierarchical structure that corresponds logically to a tree.
We assert that for a structure annotation to be consistent with a table's logical structure, there should be exactly one cell for every tree node.
We also assume that each value in the table is indexed by a unique set of keys.
We interpret this to mean that each column in the body of the table corresponds to a unique leaf node in the column header tree, and similarly that each row in the body corresponds to a unique leaf node in the row header tree (the index of a row or column can serve as a key if necessary).
These assumptions enable us to determine if a row or column header is only partially annotated and if so, to extend it to additional columns or rows, respectively.
However, to keep the precision of the algorithm high, for row headers we only attempt to infer projected row headers (PRHs, also known as \textit{projected multi-level row headers} \cite{hu2000table}, \textit{section headers} \cite{pinto2003table}, or \textit{super-rows} \cite{tengli2004learning}) and to infer cells that are in the first column of the header.
The PRHs of a table can be identified using the rule in Line \ref{algline:prh}.
Inference of the full row header is considered outside the scope of this work.

We also assume that any internal node in a header tree has at least two children.
If not, ambiguity could arise in a table's logical structure because an internal node could optionally be split into a parent node and a single child node.
The final assumptions we make are in regards to the root cause of oversegmentation in markup annotations.
We assume that cells will only be oversegmented if an oversegmentation is consistent with the table's appearance.
In practice, this means that cells with centered text will not be oversegmented in the direction of the alignment because this is likely to alter the table's appearance.
For non-centered text, we expect that when cells in either header are oversegmented, this will happen vertically, as in \cref{subfig:oversegmentation.2}, and not horizontally due to the fact that text fills horizontal space before it fills vertical space, leaving more vertical space unused.
Further, we expect that oversegmented cells in the row header will have text that is top aligned.
Finally, we expect that when projected row headers are oversegmented, this will happen horizontally, not vertically, as a projected row header already occupies only one row.

Finally, there are two additional cases that we must handle by convention.
One case is when one or more rows of blank grid cells are between a parent cell and all of its children cells in the column header.
In this case, we can choose either to merge all of the blank cells with the parent cell above it or each with the child cell below it, and we choose the convention to merge all of the blank cells with the child, which occurs in Line \ref{algline:mergecolumn}.
The final case is when a table has an blank \emph{stub head} (according to the Wang model) in its top-left corner.
In this case, the blank cells are not part of the table, so the assumptions about table structure do not suggest how they should be grouped.
We choose by convention to merge all blank cells in the same column in a blank stub head, which is consistent with the scheme in Line \ref{algline:mergecolumn}.

\begin{table*}[!t]
  \footnotesize
  \caption{Estimated measure of oversegmentation for projected row headers (PRHs) by dataset. As PRHs are only one type of cell that can be oversegmented, this is a partial survey of the total oversegmentation in these datasets.}
  \label{tab:oversegmentation}
  \centering
  \begin{tabular}{L{0.16\textwidth}R{0.1\textwidth}R{0.1\textwidth}R{0.1\textwidth}R{0.16\textwidth}R{0.16\textwidth}}
    \toprule
    \multirow{3}{*}[4pt]{\parbox{0.16\textwidth}{\textbf{Dataset}}} & \multirow{2}{*}[6pt]{\parbox{0.1\textwidth}{\begin{flushright}\textbf{Total Tables \mbox{$\textrm{Investigated}^\dagger$}}\end{flushright}}} &\multirow{2}{*}[6pt]{\parbox{0.1\textwidth}{\begin{flushright}\textbf{Total Tables \mbox{with a $\textrm{PRH}^*$}}\end{flushright}}} & \multicolumn{3}{c}{\textbf{Tables with an oversegmented PRH}} \\
    \cmidrule{4-6}
    & & &Total & \% (of total with a PRH) & \% (of total investigated) \\
    \midrule
    SciTSR & 10,431 & 342 & 54 & 15.79\% & 0.52\% \\
    PubTabNet & 422,491 & 100,159 & 58,747 & 58.65\% & 13.90\% \\
    FinTabNet & 70,028 & 25,637 & 25,348 & 98.87\% & 36.20\% \\
    \midrule
    \textbf{PubTables-1M (ours)} & 761,262 & 153,705 & 0 & \textbf{0\%} & \textbf{0\%} \\
    \midrule
    \multicolumn{6}{l}{\scriptsize $^\dagger$We exclude tables with fewer than five rows; to avoid column header rows we skip the first four rows when searching for PRHs.} \\
    \multicolumn{6}{l}{\scriptsize $^*$PRH = projected row header; these can be reliably detected in datasets without any prior row or column header annotations.} \\
  \end{tabular}
\end{table*}

\paragraph{Limitations} While the stated goal of canonicalization is applicable to any table structure annotation, we note that \cref{alg:canonicalization} is designed to achieve this specifically for the annotations in the PMCOA dataset.
Canonicalizing tables from other datasets may require additional assumptions and is considered outside the scope of this work.
Finally, it should be noted that canonicalization does not guarantee mistake-free annotations.
Remaining issues are addressed using the automated quality control procedure described next.

\paragraph{Quality control}\label{label:quality}
Because PubTables-1M is too large to be verified manually, we check for potential errors automatically and filter these from the data.
First, as tables rendered from markup should not contain overlapping rows or overlapping columns, we discard any table where this occurs, as these are likely due to mistakes introduced by the alignment process.
To filter out mistakes made both by the original annotators and our automated processing, we compare the edit distance between the non-whitespace text for every cell in the original XML annotations with the text extracted from the PDF inside the grid cell bounding box.
We filter out any tables for which the normalized edit distance between these averaged over every cell is above 0.05.
We do not force the text from each to be \textit{exactly} equal, as the PDF text can differ even when everything is annotated correctly, due to things like word wrapping, which may add hyphens that are not in the source annotations.
When the annotations do slightly differ from their corresponding PDF text, we choose to consider the PDF text to be the ground truth.
As tables with correct location information provide an unambiguous assignment of all words in the table to cells, we also compute the average fraction of overlap between each word appearing within the boundary of the table and its most overlapping grid cell, and discard tables with an average below 0.9.
Finally, we remove outliers by counting the number of objects in a table (defined in \cref{sec:model}) and removing tables with more than 100.
In all, less than 0.1\% of tables are discarded as outliers.

PubTables-1M is the first dataset that verifies annotations at the cell level and provides a measurable assurance of consistency for the ground truth.
This shows that improving the explicitness of information is valuable in part because it leads to more opportunities for catching inconsistencies and errors embedded within the data.

\paragraph{Dataset statistics and splits}\label{par:statistics} In total, PubTables-1M contains 947,642 tables for TSR, of which 52.7\% are complex (have at least one spanning cell).
Prior to canonicalization, only 40.1\% of the tables in the set were considered complex by the original annotators.
Canonicalization adjusts the annotations in some way for 34.7\% of tables, or 65.8\% of complex tables.

To further assess the impact on oversegmentation, we compare our final dataset with other datasets in \cref{tab:oversegmentation}.
Precisely measuring the amount of oversegmentation in a dataset requires annotations for the row and column headers.
But because other datasets lack these, we instead measure just oversegmented projected row headers (PRHs).
PRHs can be detected reliably without explicit annotations using the rule in Line \ref{algline:prh}.
To account for missing column header annotations, we do not start looking for PRHs until at least the fifth row, which assumes the column header occupies at most four rows, and we simply exclude any tables that have fewer than five rows.
In case there are un-annotated footers, we also do not count any detected PRHs that are the last rows of the table.
A detected PRH is oversegmented if its row contains a blank cell.
As can be seen, canonicalization eliminates a significant source of oversegmentation, and thus ambiguity, that is present in other datasets.
Interesting to note, FinTabNet nearly always oversegments projected row headers.
While self-consistent, this widespread oversegmentation contradicts the logical structure of the table and causes potential issues with combining this dataset with others that annotate these rows differently.

We split PubTables-1M randomly into train, validation, and test sets at the document level using an 80/10/10 split.
For TSR, this results in 758,849 tables for training; 94,959 for validation; and 93,834 for testing.
For TD, there are 460,589 fully-annotated pages containing tables for training; 57,591 for validation; and 57,125 for testing.
An example page and table annotation for TD is shown in \cref{fig:table_extraction_tasks}.
Note that tables that span multiple pages are considered outside the scope of this work.

\section{Proposed Model}\label{sec:model}

We model all three tasks of TD, TSR, and FA as object detection with images as input.
For TD, we use two object classes: \textit{table} and \textit{table rotated}.
The \textit{table rotated} class corresponds to tables that are rotated counterclockwise 90 degrees.

\begin{figure}[]
  \centering
  \includegraphics[width=0.95\linewidth]{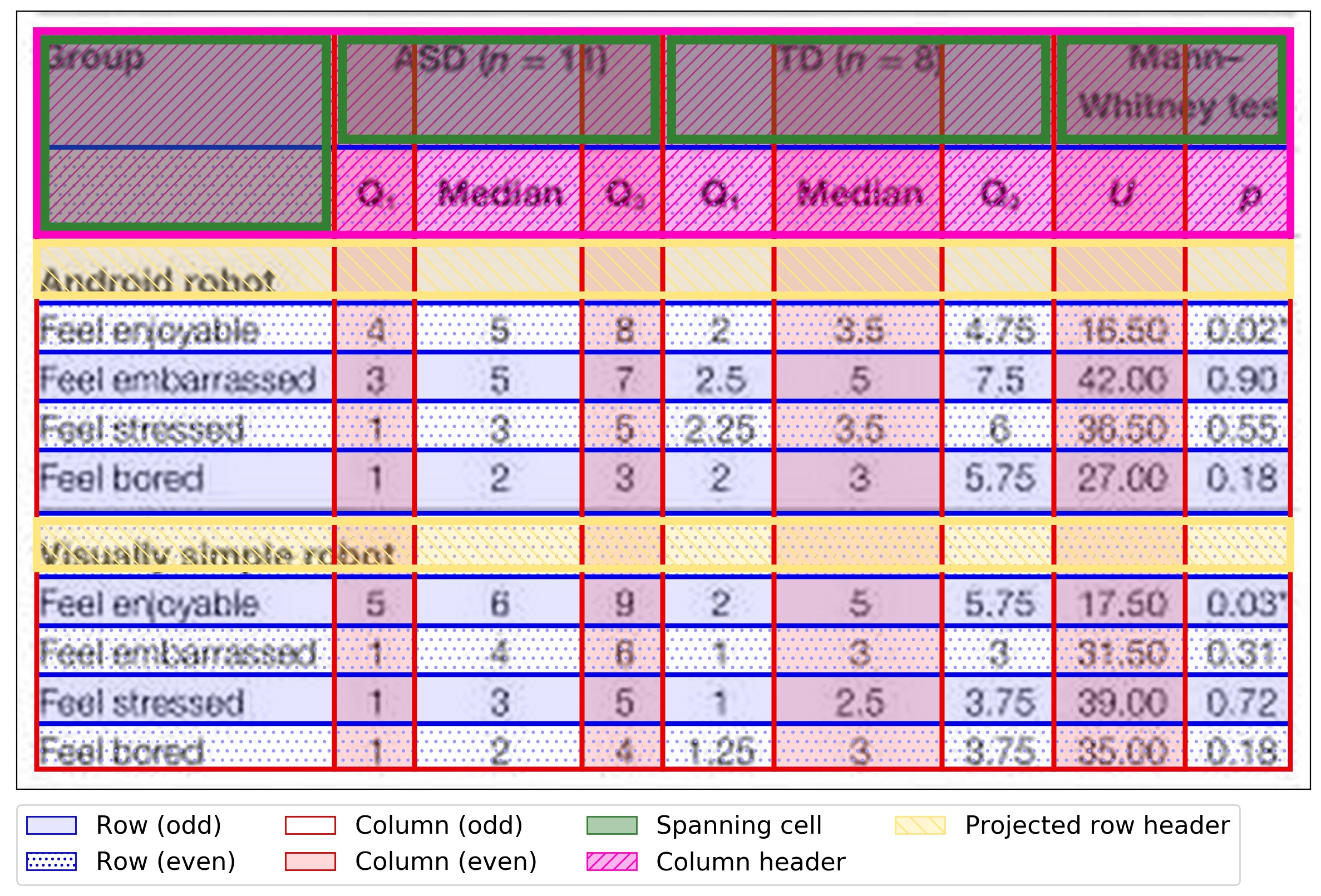}
  \caption{An example table with dilated bounding box annotations for different object classes for jointly modeling table structure recognition and functional analysis.}
  \label{fig:object_classes}
\end{figure}

\paragraph{TSR and FA model} We use a novel approach that models TSR and FA jointly using six object classes: \textit{table}, \textit{table column}, \textit{table row}, \textit{table column header}, \textit{table projected row header}, and \textit{table spanning cell}.
We illustrate these classes in \cref{fig:object_classes}.
The intersection of each pair of \textit{table column} and \textit{table row} objects can be considered to form a seventh implicit class, \textit{table grid cell}.
These objects model a table's hierarchical structure through physical overlap.

For the TSR and FA model, we use bounding boxes that are \textit{dilated}.
To create dilated bounding boxes, for each pair of adjacent rows and each pair of adjacent columns, we expand their boundaries until they meet halfway, which fills the empty space in between them.
Similarly we expand the objects from the other classes so their boundaries match the adjustments made to the rows and columns they occupy.
After, there are no gaps or overlap between rows, between columns, or between cells.

To demonstrate the proposed dataset and the object detection modeling approach, we apply the Detection Transformer (DETR) \cite{carion2020end} to all three TE tasks.
We train one DETR model for TD and one DETR model for both TSR and FA.
For comparison, we also train a Faster R-CNN \cite{ren2015faster} model for the same tasks.
All models use a ResNet-18 backbone pre-trained on ImageNet with the first few layers frozen.
We avoid custom engineering the models and training procedures for each task, using default settings wherever possible to allow the data to drive the result.

\section{Experiments}\label{sec:experiments}

In this section, we report the results of training the proposed models on data derived from PubTables-1M.
For TD, we train two models: DETR and Faster R-CNN.
We report the results in \cref{detection-results}.
For table detection, DETR slightly outperforms Faster R-CNN on $\textrm{AP}_\textrm{50}$ but significantly outperforms on AP.
We interpret this to mean that while both models are able to learn to detect tables, DETR precisely localizes tables much better than Faster R-CNN.

\begin{table}[!t]
  \footnotesize
  \caption{Test performance of models on PubTables-1M using object detection metrics.}
  \label{detection-results}
  \centering
  \begin{tabular}{lccccc}
    \toprule
    \textbf{Task} & \textbf{Model} & \textbf{AP} & $\textbf{AP}_\textbf{50}$ & $\textbf{AP}_\textbf{75}$ & \textbf{AR}\\
    \midrule
    TD & Faster R-CNN & 0.825 & 0.985 & 0.927 & 0.866 \\
    & DETR & \textbf{0.966} & \textbf{0.995} & \textbf{0.988} & \textbf{0.981} \\
    \midrule
    TSR + FA & Faster R-CNN & 0.722 & 0.815 & 0.785 & 0.762 \\
    & DETR & \textbf{0.912} & \textbf{0.971} & \textbf{0.948} & \textbf{0.942} \\
    \midrule
  \end{tabular}
\end{table}

\begin{table*}
\footnotesize
  \caption{Test performance of the TSR + FA models on PubTables-1M on TSR metrics.}
  \label{proposed-results}
  \centering
  \begin{tabular}{l|l|l|cccccccc}
    \toprule
    \textbf{Test Data} & \textbf{Model} & \textbf{Table Category} & $\textbf{Acc}_\textbf{Cont}$ & $\textbf{GriTS}_\textbf{Top}$ & $\textbf{GriTS}_\textbf{Cont}$ & $\textbf{GriTS}_\textbf{Loc}$ & $\textbf{Adj}_\textbf{Cont}$ \\
    \midrule
    Non-Canonical & DETR-NC & Simple & 0.8678 & 0.9872 & 0.9859 & 0.9821 & 0.9801 \\
    & & Complex & 0.5360 & 0.9600 & 0.9618 & 0.9444 & 0.9505 \\
    & & All & 0.7336 & 0.9762 & 0.9761 & 0.9668 & 0.9681 \\
   \midrule
    Canonical & DETR-NC & Simple & 0.9349 & 0.9933 & 0.9920 & 0.9900 & 0.9865 \\
    & & Complex & 0.2712 & 0.9257 & 0.9290 & 0.9044 & 0.9162 \\
    & & All & 0.5851 & 0.9576 & 0.9588 & 0.9449 & 0.9494 \\
    \cmidrule{2-8}
    & Faster R-CNN & Simple & 0.0867 & 0.8682 & 0.8571 & 0.6869 & 0.8024 \\
    & & Complex & 0.1193 & 0.8556 & 0.8507 & 0.7518 & 0.7734 \\
    & & All & 0.1039 & 0.8616 & 0.8538 & 0.7211 & 0.7871 \\
\cmidrule{2-8}
    & DETR & Simple & \textbf{0.9468} & \textbf{0.9949} & \textbf{0.9938} & \textbf{0.9922} & \textbf{0.9893} \\
    & & Complex & \textbf{0.6944} & \textbf{0.9752} & \textbf{0.9763} & \textbf{0.9654} & \textbf{0.9667} \\
    & & All & \textbf{0.8138} & \textbf{0.9845} & \textbf{0.9846} & \textbf{0.9781} & \textbf{0.9774} \\
    \midrule
  \end{tabular}
\end{table*}

For TSR and FA, we train three models: Faster R-CNN and DETR on the canonicalized data, and DETR on the original, non-canonical (NC) annotations (DETR-NC).
We report the results using object detection metrics for the models trained on canonical data in \cref{detection-results}, which measures performance jointly on TSR and FA, and report results for all models using TSR-only metrics in \cref{proposed-results}.
For TSR only, we also evaluate DETR-NC on both the canonical and the original non-canonical test data.

For assessing TSR performance, we report the table content accuracy metric ($\textrm{Acc}_\textrm{Cont}$), which is the percentage of tables whose text content matches the ground truth exactly for every cell, as well as several metrics for partial table correctness, which use different strategies to give credit for correct cells when not all cells are correct.
For partial correctness, we use the F-score of the standard adjacent cell content metric \cite{gobel2012methodology} and the recently proposed GriTS metrics \cite{smock2021grits}.
GriTS metrics have the form,

\begin{align}
\label{eq:grits}
\mathrm{GriTS}_f(\mathbf{A}, \mathbf{B}) = \frac{2 \cdot \sum_{i,j} f(\mathbf{\tilde{A}}_{i,j}, \mathbf{\tilde{B}}_{i,j})} {{|\mathbf{A}|} + {|\mathbf{B}|}},
\end{align}

\noindent which can also be interpreted as an F-score.
GriTS represents the ground truth and predicted tables as matrices, $\mathbf{A}$ and $\mathbf{B}$, and computes a similarity score between the most similar substructures \cite{amir2008generalized} of these matrices, $\mathbf{\tilde{A}}$ and $\mathbf{\tilde{B}}$, where a substructure is defined as a selection of $m$ rows and $n$ columns from the matrix.
Compared to other metrics for TSR, this formulation better captures the two-dimensional structure and ordering of cells of a table when comparing tables.
Further, GriTS enables TSR to be assessed from multiple perspectives within the same formulation, with $\textrm{GriTS}_\textrm{Top}$ measuring cell topology recognition, $\textrm{GriTS}_\textrm{Cont}$ measuring cell content recognition, and $\textrm{GriTS}_\textrm{Loc}$ measuring cell location recognition.

As can be seen in \cref{proposed-results}, DETR trained on the canonical data produces strong results for TSR and FA, outperforming the other models when evaluated on all tables.
Comparing DETR-NC evaluated on NC ground truth versus DETR evaluated on canonical ground truth, we see that using canonical data improves performance on the metrics across all table types.
This is even more apparent for the table accuracy metric, for which the use of canonical data is responsible for a jump in performance from 0.5360 to 0.6944 for complex tables.

To consider the positive impact that canonicalization has just on facilitating a more reliable evaluation, we compare DETR-NC evaluated on canonical data versus NC data.
Even though it is trained on NC data, the metrics for simple tables are much higher when DETR-NC is evaluated on canonical data (0.9349 accuracy) than when it is evaluted on NC data (0.8678 accuracy).
This shows even more clearly that the canonical data is less noisy and contributes to a more reliable evaluation.

Overall the results confirm that canonical data significantly improves performance for TSR models.
Even setting aside the argument that non-canonical data is noisier and more ambiguous, it is also a \emph{different} way of annotating the data, using oversegmentation, which is less useful because it does not correspond to a table's true logical structure.
This difference is apparent when we compare DETR-NC versus DETR, both evaluated on canonical test data.
DETR-NC performs much worse on complex tables due in large part to the inconsistent scheme it learns for understanding the spanning cells in their headers.

\section{Conclusion}

In this paper, we introduced a new dataset, PubTables-1M, for table extraction in unstructured documents.
PubTables-1M address the challenge of creating complete, reliable ground truth at scale for table structure recognition.
We called attention to the problem that oversegmentation in markup annotations leads to ambiguous ground truth in crowd-sourced datasets, and proposed a novel \emph{canonicalization} procedure to address this.
We demonstrated that the proposed improvements to the ground truth data have a significant positive impact on model performance.
Finally, we adopted DETR for all three table extraction tasks and showed for the first time that it is possible to achieve state-of-the-art performance within a  standard object detection framework without the need for any special customization for these tasks.
While we do not believe this work raises any issues regarding negative impacts to society, we welcome a discussion on any potential impacts raised by others.

\section{Future Work}

In the future, we hope to expand the proposed methods and canonicalization beyond scientific articles to data from additional domains, such as tables in financial documents.
We also hope to address the open challenge of accurately annotating row headers in large-scale datasets, which will enable even more complete solutions for table extraction.
Finally, table extraction is often just one stage in larger pipelines for document understanding and information retrieval, and developing end-to-end systems in these areas is an important future direction with its own challenges.
We hope that releasing a large pool of detailed table annotations from the PMCOA corpus in particular can further progress in this area.
\section{Acknowledgments}
We would like to thank Pramod Sharma, Natalia Larios Delgado, Joseph N. Wilson, Mandar Dixit, John Corring, and Ching Pui WAN for helpful discussions and feedback while preparing this manuscript.

{\small
\bibliographystyle{ieee_fullname}
\bibliography{paper}
}

\section{Appendix}

\subsection{Training Details}

For both DETR models, we use a ResNet-18 backbone, six layers in the encoder, and six layers in the decoder.
For TD, we use 15 object queries, and for TSR and FA we use 125 object queries, each chosen to be slightly more than the maximum number of objects in each set's training samples.
Besides this, we use the same default architecture settings for each model.

All of the experiments are performed using a single NVidia Tesla V100 GPU.
We initialize the models with weights pre-trained on ImageNet and train each model for 20 epochs using all default hyperparameters and training settings except for those we note here.
For both models, we use a learning rate drop of 1 and gamma of 0.9.
For the TSR and FA model, we also use an initial learning rate of 0.00005 and a no-object class weight of 0.4.
We limited hyperparameter tuning to one short experiment to determine the initial learning rate. 
We ran training experiments with three different initial learning rates of 0.0002, 0.0001, 0.00005 and chose to use the learning rate for each model that had the best performance on the validation set after one epoch of training.

We use no custom components, losses, or procedures for training the model, other than standard data augmentations, such as random cropping and resizing.
To create training data for the TD model, we render the PDF pages to images with a maximum length of 1000 pixels and appropriately scale the bounding boxes for the objects to image coordinates.
For TSR and FA, we first render the page containing the table as an image with a maximum length of 1000 pixels, scale and pad the table bounding box with an additional 30 pixels on all sides (or fewer on a side if there are less than 30 pixels available on that side), and crop the image to this bounding box.
The padding enables more variation in training through cropping augmentations.

\subsection{Inference}

While the model is trained entirely as an object detector, at inference time an additional step is required to convert from objects output by the model to a structured table.
Objects in our model form hierarchical relationships, where a parent-child relationship between objects is indicated by physical containment, or overlap.
For details on the full set of relationships in the hierarchy, please see the code.
One thing we mention here is that according to this hierarchy no object can have two parents of the same object class.
Thereforem, a \emph{conflict} occurs when two parent objects of the same class both overlap with a potential child object for that class.
For example, two spanning cell objects are considered to conflict if they both overlap with the same grid cell.
The definition of overlap can vary by object class but is generally when 50\% of a child object's bounding box by area overlaps with a parent object's bounding box.

The ground truth data is conflict-free, so at training time the model learns to produce output that is also free of conflicts.
However, to address any conflicts produced by the model strictly at inference time once training has finished, we add to the models a simple \textit{conflict resolution} step.
The conflict resolution step involves suppressing objects or adjusting their bounding boxes to eliminate conflict between objects of the same class.
This is similar to non-maxima suppression, except overlap between objects is defined in terms of children objects instead of pixels.
Once the objects are conflict-free, they are converted to a logical table.

\subsection{Scoring}

Lastly, we note that for the sake of evaluation, once the text that occurs within each grid cell is extracted, we retroactively adjust the vertical extent of the row bounding boxes and the horizontal extent of the column bounding boxes to tightly wrap the text they contain.
Note that cells contain the same text content before and after this tightening.
We then use these adjusted bounding boxes for cell location scoring purposes.
This adjustment is necessary to properly score because we train the model using dilated bounding boxes, which expand the boundaries between rows and between columns to fill the gaps between them, thereby containing extraneous padding.
This extraneous padding is useful for learning but would unfairly penalize the model if included at scoring time, therefore we consider the tightened cells to be the actual cell locations output by the model and score with these.

\end{document}